\documentclass[11pt,a4paper]{article}
\usepackage[hyperref]{emnlp2018}
\usepackage{times}
\usepackage{latexsym}
\usepackage{graphicx}
\usepackage{todonotes}

\usepackage{url}

\usepackage{algorithm2e}

\usepackage{amsmath}
\DeclareMathOperator*{\argmax}{argmax}

\aclfinalcopy 

\title{Supervised Machine Learning for Extractive Query Based Summarisation of Biomedical Data}

\author{Mandeep Kaur \\
  Macquarie University \\
  Sydney, Australia \\
  {\tt mandeep-kaur.mandeep-kaur@} \\
  {\tt students.mq.edu.au} \\\And
  Diego Moll\'a \\
  Macquarie University \\
  Sydney, Australia \\
  {\tt diego.molla-aliod@mq.edu.au} \\}

\date{}

\begin{document}
\maketitle
\begin{abstract}
The automation of text summarisation of biomedical publications is a pressing need due to the plethora of information available on-line. This paper explores the impact of several supervised machine learning approaches for extracting multi-document summaries for given queries. In particular, we compare classification and regression approaches for query-based extractive summarisation using data provided by the BioASQ Challenge. We tackled the problem of annotating sentences for training classification systems and show that a simple annotation approach outperforms regression-based summarisation.
\end{abstract}

\section{Introduction}

Text summarisation is a task of abridgement full text into a compact version while preserving the crucial information of the original text that is relevant to a user. The continuous increase of volume of digital text over the internet has reached such tremendous magnitude that a plethora of on-line text is available in regard to a topic. Consequently, manual skimming of text faces paramount obstacles like information overload \cite{Das:2007}. 
This problem is particularly important for medical practitioners who need to analyse all the relevant information to diagnose and determine the best course of action for a particular patient. 
For example, there are cases in which medical practitioners fail to pursue answers to their queries \cite{Ely:2005}. Moreover, manually searching the information is an extremely time-consuming and expensive task. Therefore, there is a strong motivation for building text processing systems that can automate some of the processes involved in this practice. 

Our focus is to perform query-focused summarisation, also known as user-focused summarisation, of biomedical publications, by extracting and summarising the content relevant to the query given by the practitioner. The extraction system used in our experiments takes into account a specific query written as a question in plain English and tries to identify the information within a set of retrieved documents that is relevant to the query. Motivated by the success of machine learning in automatic text summarisation, we address the task of automatic query-based summarisation of biomedical text by using supervised machine learning techniques. We generate summaries by identifying the most significant content from the input text within the context of a query and generating a final summary by utilising that content.

In addition, this research also deals with a burning issue of availability of annotated corpora for supervised learning. In computational linguistics, labelled corpora are used to train machine learning algorithms and assess the performance of automatic summarisation methods. The employment of annotated corpora to the field of summarisation dates back to the late 1960s. These annotations typically consist of human-produced summaries, and it is not trivial to determine how to convert this information into the specific annotations required for supervised machine learning approaches to summarisation. 
Getting this data manually labelled is quite expensive and time-consuming; automatic annotation of data is still an active research question.

The contributions of this paper include: 
\begin{enumerate}
\item A comparison of supervised approaches to query-focused extractive text summarisation of biomedical data.
\item A comparison of annotation approaches for classification-based approaches to query-focused extractive summarisation of biomedical data.
\end{enumerate} 

The rest of the paper is organised as follows. Section~\ref{sec:related} provides a brief review of related work on the topic of extractive summarisation, with references to systems using biomedical text. Section~\ref{sec:data} discusses the BioASQ Challenge, whose data are used in our experiments, and how it relates to query-focused summarisation. Section~\ref{sec:model} presents the details of our summarisation framework. Section~\ref{sec:annotation} discusses various annotation approaches used to train classifiers for supervised machine learning. Section~\ref{sec:results} illustrates the results of our experiments for regression and classification approaches, along with an analysis of the output from our classification models using different annotation approaches. Finally, Section~\ref{sec:conclusions} concludes the paper with remarks on our future direction.

\section{Related Work}\label{sec:related}


Text summarisation has a rich background of research algorithms starting form late 1950's. The earliest works on text summarisation used sentence extraction as a primary component of a text summarisation system and the classic extractive approaches applied to extract summaries used statistical features for selecting significant content from the source text. The text features utilised by these approaches were based on bag-of-words (BOW) approaches. BOW models including word frequency and tf-idf are the most frequently used methods to discover the important content \cite{Wu:2008}. More recently, word embeddings generated by deep learning approaches have also been shown to be useful for text summarisation \cite{Malakasiotis2015,Molla2017}.

In recent years, the main focus of research in the summarisation field has been directed towards the application of machine learning to generate better summaries. 
Popular features such as multiple words, noun phrases, main verbs, named entities and word embeddings \cite{Barzilay:1997,Filatova:2004,Harabagiu:2002,Malakasiotis2015,Molla2017} have been heavily exploited for summarisation.

In contrast to other domains, research on automatic text processing in the medical domain is still very much in its infancy. In the recent past, there has been steady ongoing research in biomedical text processing \cite{Zweigenbaum:2007}. Factors such as the requirement of large volume of data, highly complex domain-specific terminologies  and domain-specific format, and typology of questions \cite{Athenikos2010a} makes it complex to process biomedical text.
Most of the researchers working on summarisation for the medical domain apply the same kinds of techniques developed in other domains. 

Three main supervised machine learning approaches have been used for text summarisation: classification, regression, and learning to rank.

\paragraph{Classification:} 
The concept of summarising text by using supervised classification approaches was pioneered by \citet{Kupiec:1995}. They categorised each sentence as worthy of extraction or not by a classification function, using a Na\"ive Bayes classifier. In this classification approach the sentences are treated individually. 
At first, most machine learning systems assumed feature independence and relied on Na\"ive Bayes methods \cite{Das:2007}. However, later models shifted the focus towards breaking the assumption that features are independent of each other \cite{Lin:1999}.

Classification approaches have also been applied for summarisation of biomedical text.
A work proposed by \citet{Chuang:2000} used decision trees and Na\"ive Bayes classifiers to train the summariser to extract important sentence segments based on feature vectors in order to generate a final summary.
Other work by \citet{Sarkar:2009} and \citet{Sarkar:2011} applied classification techniques to extractive summarisation by classifying individual sentences. The features used were term frequency, sentence similarity to document title, position of sentence, presence of domain specific cue phrases, presence of novel terms, and sentence length.

\paragraph{Regression:} Regression approaches for summarisation try to fit the predicted score of a sentence as close as possible to the target score instead of labelling the sentences. An early work using regression for summarisation is by \citet{Ouyang:2011} using support vector regression (SVR). Support vector regression (SVR) has also been used in conjunction with other techniques like integer linear programming (ILP) for generating summaries \cite{Galanis:2012} and has achieved state-of-the-art results in comparison to other competitive extractive summarisers.

A system named FastSum \cite{Schilder:2008} used regression SVM for training their data set by using the least computationally expensive NLP techniques to generate the summary. The system used a set of clusters as input data and simple pre-processing was performed on the sentences. A comparison of this system with MEAD \cite{Radev:2000} showed that it is more than 4 times faster than MEAD.

Some of the recent work on biomedical data \cite{Malakasiotis2015} used BioASQ data which is the data used in this paper. As in this paper, their work addressed the task of multi-document query focused summarisation. They used SVR to assign relevance scores to the sentences of the given relevant abstracts, and an alternative greedy strategy to select the most relevant sentences avoiding redundant ones. 

A system by \citet{Molla2017} also experimented using BioASQ data in conjunction with SVR. The feature set used was  based on \citet{Malakasiotis2015}. In addition to SVR, \citet{Molla2017} used other regression approaches with deep learning architectures including convolutional neural networks (CNNs) and long-short term memory networks (LSTMs).  

\paragraph{Learning to rank:} Learning to rank transforms the task into a simple problem of ranking extracts from an original text. Given sentences with labelled importance scores, it is possible to get learning to rank models to train a model capable of assigning high rank to the most important sentences. 

Ranking SVMs are the most commonly used approaches for learning to rank. When comparing SVMs and ranking SVMs to model the relevance of sentences to queries, \citet{Wang:2007} show that ranking SVMs outperform standard SVMs on a small test collection. Learning to rank has also been applied to the summarisation of XML documents with a goal of learning how to best combine the sentence features such that within each document, summary sentences get higher scores than non-summary ones \cite{Amini:2007}. 

Another significant work done in this category uses ranking SVM to combine features for extractive query focused multi-document summarisation \cite{Shen2011}. In order to do that, a graph-based method was proposed for training data generation by utilizing the sentence relationships and a cost sensitive loss was introduced to improve the robustness of learning. The method outperformed the baseline strategies.

We are not aware of any work on biomedical summarisation using learning to rank techniques.

\section{The BioASQ Challenge}\label{sec:data}

We utilised a biomedical corpus provided by the BioASQ Challenge\footnote{\url{http://bioasq.org/}}. The BioASQ Challenge organises shared tasks on aspects related to biomedical semantic indexing and question answering \cite{Tsatsaronis:2015}. One of the tasks, Task~B, focuses on question answering, and Phase ~B of Task~B asks participants to respond to a query by providing ``exact answers'' and ``ideal answers''. Whereas the exact answers are the usual output of a factoid question answering system, the ideal answers contain additional text such as explanations and justifications, and can be viewed as examples of query-focused summarisation. Figure~\ref{fig:bioasq} shows an example of a question, its exact answer, and its ideal answer, as provided in the training set of BioASQ~5b.
\begin{figure}
\centering
\begin{description}
\item[Query:] Name synonym of Acrokeratosis paraneoplastica.
\item[Exact answer:] Bazex syndrome.
\item[Ideal answer:] Acrokeratosis paraneoplastic (Bazex syndrome) is a rare, but distinctive paraneoplastic dermatosis characterized by erythematosquamous lesions located at the acral sites and is most commonly associated with carcinomas of the upper aerodigestive tract.
\end{description}
\caption{Example of query, exact answer, and ideal answer from the BioASQ~5b Phase~B shared task.\label{fig:bioasq}}
\end{figure}

In the BioASQ data set each question contains, among other information, the text of the question, the question type, and a list of source documents. The list of documents has been extracted manually by annotators and are relevant to the query. They can be viewed as the ideal output of a text retrieval system and are used as the input data of our experiments. The training data set contains a total of 1306 questions.

\section{Summarisation Model}\label{sec:model}

\begin{figure}
\centering
\includegraphics[width=0.4\linewidth]{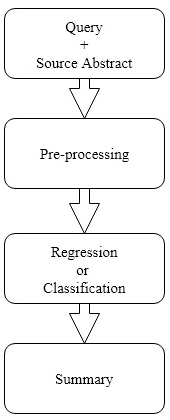}
\caption{The overall summarisation model.\label{fig:model}}
\end{figure}

Our system performs query-focused extractive summarisation of biomedical data, and our model is trained with data from the BioASQ~5b Challenge.
We follow a three-stage summarisation model for the generation of the summaries. In the first stage, the question and input text are pre-processed and transformed to an intermediate representation. In the second stage, each sentence in the input is assigned an importance score or label depending on the approach applied. Finally, in the final stage, the $n$ most highly ranked sentences are selected to generate a summary. Figure~\ref{fig:model} outlines the summarisation model.

\subsection{Pre-processing}

Pre-processing refers to the first stage of the model. First, the data are partitioned into training and testing using 10-fold cross validation. After partitioning the data, the sentences and questions are vectorised by computing the tf-idf of their words.

We also incorporate a technique that compares sentences with the associated queries. In particular, we compute the cosine similarity of each candidate ($S_i$) sentence with the associated query ($Q_i$), using the tf-idf vector representations for each:
$$
Sim(S_i,Q_i) =\frac{S_i \cdot Q_i}{\lVert S_i\rVert\, \lVert Q_i\rVert} 
$$

\subsection{Approaches for Extracting Summaries}

Regression and classification-based techniques are used for generating a summary for a given query. To enable the comparison of all techniques, we have used a common feature set. In our case the feature set used is: 

\begin{enumerate}
\item tf-idf vector of the candidate sentence.
\item Cosine similarity between the tf-idf vector of the question and the tf-idf vector of the candidate sentence. 
\end{enumerate}

Since the intent of this work is to compare the performance of regression and classification approaches, and not to obtain the best possible results, the feature set used is fairly simple and is commonly used on the most popular supervised approaches for query-based extractive summarisation.

For the regression approaches, each sentence of the training data is annotated with the F1 ROUGE-SU4 score of the sentence compared to the target summary. 
ROUGE-SU4 considers skip bigrams with a maximum distance of 4 words between the words of each skip bigram \cite{Lin:2004}. This measure has also been found to correlate well with human judgements in extractive summarisation. Other systems have used ROUGE for annotating data and its application has been proved useful, e.g the system by \citet{Galanis:2012,Peyrard2016}. We use Support Vector Regression (SVR), which has performed well in past regression approaches to summarisation.

For the classification approaches, we use the standard two-class labelling approach where class~1 indicates sentences that are selected for the final summary, and class~0 indicates sentences that are not selected. 
We use Support Vector Machine (SVM), which has performed well in many other classification problems.

\section{Data Annotation for Classification}\label{sec:annotation}

Supervised machine learning requires annotated training data to generate summaries. Often the summary annotations consist of sample reference summaries, but it is not straightforward to translate this information into the target labels~1 and~0 for classification. Although many researchers attempted to tackle this issue by manually selecting the summary-worthy sentences for their experiments \cite{Ulrich:2008}, manual annotation consumes a considerable amount of time. 

We have experimented with several approaches to determine when to assign a label~1 or~0 to an input sentence for the training procedure. As mentioned above, the inherent annotation of the sentences for the regression approach is based on their ROUGE score. Whereas it is straightforward to use ROUGE for the regression approach, we need to convert the ROUGE score into a binary value for classification. We experimented with various thresholds, and compared with a more complex approach based on \citet{Marcu:1999}'s work.

\subsection{ROUGE Annotation with Thresholds}

We tried two thresholds to define the labels for both the summary and the non-summary classes so that, if the ROUGE-SU4 score of the sentence is above the threshold, the sentence is labelled~1. Otherwise the sentence is labelled~0. This is done for every sentence associated with a query.

Firstly, we experimented by labelling the three highest SU4 scoring sentences as summary (i.e. label~1) for each query in the data. Secondly, we tried a threshold of~0.1. We labelled the sentence as~1 if its SU4 score is higher than~0.1 and labelled the rest as~0. 

\subsection{Marcu Annotation}

In addition to the above-mentioned ROUGE annotation approaches, we also experimented with a greedy approach proposed by \citet{Marcu:1999} that we call the Marcu annotation. The motivation behind using this approach for our experiments is that it takes into account the similarity between the target abstract and the entire set of sentences selected for the summary.

This method, instead of selecting sentences which are identical to those in the abstract, eliminates sentences which do not appear to be similar to ones in the abstract. The rationale of the methodology is that, if the similarity between the document and its target abstract does not decrease when a sentence is removed from the document, then we can say that the sentence is not relevant to the target abstract \cite{Marcu:1999}. This elimination process continues while the similarity does not decrease as we remove sentences.

The original algorithm by \citet{Marcu:1999} is divided into two parts: generating the core extract and cleaning-up the core extract. The first part of the algorithm results in an extract through which important sentences in the text can be identified and annotated. In the second part, some cosmetic procedures are performed to the generated extract. In this second clean-up step Marcu employed some heuristics to further reduce the set of sentences.

We only implemented the first part of the algorithm. There are two reasons for not implementing the second part of the algorithm. Firstly, some of the heuristics require knowledge of the rhetorical structure of the source to be able to apply them. This information was not available, and could not be easily obtained. In addition, for some of the heuristics, the details were insufficient to know exactly how to implement them.

Algorithm~\ref{algo} shows the algorithm for generating the extract. The input to the algorithm is a reference abstract and input text to summarise. In step~1, the input text is broken into sentences. Step~2 then pre-processes the abstract and text. Pre-processing involves tokenising all the information into words and then performing stemming and removing stop words. We use NLTK for steps~1 and~2 in contrast to \citet{Marcu:1999}'s approach, who used a shallow clause boundary and discourse marker identification (CB-DM-I) algorithm for this task. This algorithm is more complex and considers the information related to various textual units to perform pre-processing.

Initially, we assume the extract to be the whole text (step~3 in Algorithm~\ref{algo}). 

Steps~4 and~5 can be explained as follows: If we delete from $E$ a sentence $S$ that is totally distinct from the abstract $A$, we obtain a new extract $E\backslash S$ whose similarity with $A$ is higher than that of $E$. We therefore apply a greedy approach and repeatedly delete sentences from $E$ so that at each step the resulting extract has maximum similarity with the abstract. We eventually reach a state where we can no longer delete sentences without decreasing the similarity of $E$ with the abstract. The resulting $E$ at this stage is considered the extract that we are looking for. 
\begin{algorithm}
\DontPrintSemicolon
\KwData{\\
Abstract ($A$):  The reference summary.\\ 
Text ($T$): Input text to summarise.\\
}
\KwResult{\\
Extract ($E$): A set of sentences from text which has maximum similarity to abstract}
\vspace{1ex}
    \nl $T_1, \cdots\ T_n$ = sentences from $T$\;
    \nl Stem and delete stop words from $A, T_1, \cdots\ T_n$\;
    \nl $E$ = T\;
    \nl $S = \argmax_{S'\in E}Sim(E\backslash S', A)$\;
    \nl \While{$Sim(E,A) < Sim(E\backslash S, A)$}{
                             $E = E\backslash S$\;
                             $S = \argmax_{S'\in E}Sim(E\backslash S', A)$\;}
\caption{Marcu's greedy approach for the generation of a core extract.\label{algo}}
\end{algorithm}

The similarity operator $Sim(X,Y)$ is the cosine similarity between the tf-idf of $X$ and $Y$.

\section{Evaluation and Results}\label{sec:results}



We evaluated all our approaches automatically using the ROUGE evaluation tool \cite{Lin:2004}. Our system-generated summaries are all evaluated by comparing them with the associated gold standard summaries which are the BioASQ ideal answers in our case.





Figure~\ref{fig:all_approaches} shows the results of the regression and the following classification approaches described in Section~\ref{sec:annotation}:
\begin{figure}
\centering
\includegraphics[width=\linewidth]{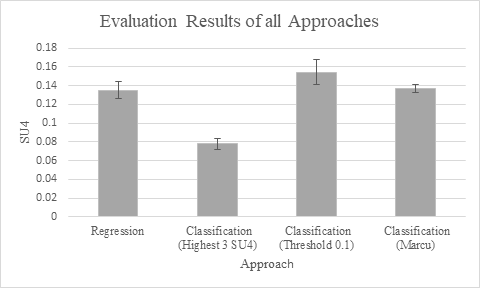}
\caption{Comparison of the results of the regression and three classification approaches. The results show the mean of 10-fold cross-validation, and the error bars show the standard deviation.\label{fig:all_approaches}}
\end{figure}

\begin{enumerate}
\item Label with~1 the three sentences with highest ROUGE score per question.
\item Label with~1 all sentences with ROUGE score higher than 0.1.
\item Label with~1 the sentences annotated according to Algorithm~\ref{algo}.
\end{enumerate}

\subsection{Regression Versus Classification}

To produce comparable results, we kept preprocessing, feature extraction and number of sentences (3 sentences) in the final summary constant. The same data partition into training and testing was used in all cases. 

\begin{figure}
\centering
\includegraphics[width=\linewidth]{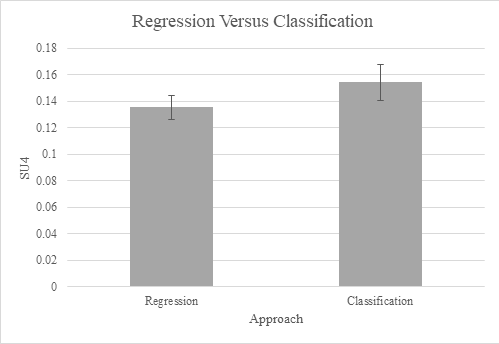}
\caption{Comparison of classification (with 0.1 threshold) and regression according to their ROUGE-SU4 (error bars refer to standard deviation) in 10-fold cross-validation.\label{fig:clas-reg}
}
\end{figure}

Figure~\ref{fig:clas-reg} compares the F1 ROUGE-SU4 scores of regression and the best classification approach. We can observe that the average SU4 score of the classification approach is higher than the score of the regression approach. The classification approach mentioned in Figure~\ref{fig:clas-reg} is the one with threshold 0.1. The standard deviation for both approaches is indicated by the error bars.

\begin{figure}
\centering
\includegraphics[width=\linewidth]{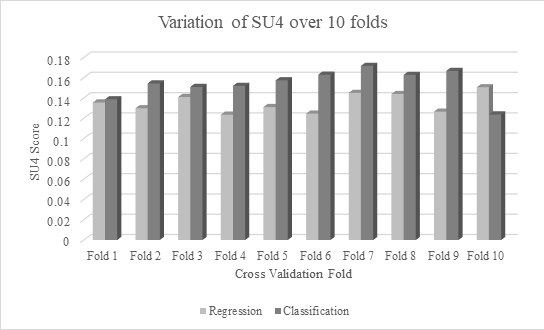}
\caption{ROUGE SU4 variation over 10-Fold cross-validation for classification and regression.\label{fig:10folds}
}
\end{figure}

To have a more precise evaluation, we analyse the variation of SU4 at each cross-validation fold for each approach to see whether classification is performing better than regression at every fold of cross-validation. In Figure~\ref{fig:10folds}, the variation of the SU4 score over each of the 10 folds for both techniques is shown and classification SU4 can be seen on the higher side for all the folds except for the last one.

\subsection{Comparing Annotation Approaches}

\begin{figure}
\centering
\includegraphics[width=\linewidth]{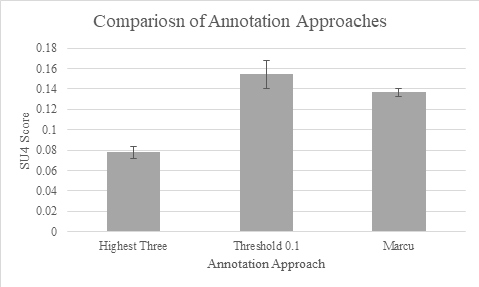}
\caption{Comparison of various annotation approaches (error bars refers to standard deviation)\label{fig:annotations}
}
\end{figure}

Figure~\ref{fig:annotations} shows F1 ROUGE-SU4 scores of all of the classification approaches: (i) using three sentences with highest SU4 as summary class, (ii) use threshold 0.1, and (iii) use the approach based on \citet{Marcu:1999}'s work.  

The second approach (i.e. with threshold 0.1) can be seen as outperforming all the other approaches. In contrast, the first approach produces the lowest SU4 score among all the three. Whereas Marcu's approach is better than the approach with the highest three, it is outperformed by the approach with threshold of 0.1. The standard deviations for all of the approaches through 10-fold cross-validation are also presented as error bars in Figure~\ref{fig:annotations}. 

\subsection{Comparison with Ouyang et al.}

A similar work performed by \citet{Ouyang:2011} reported better results for regression than for classification in their experiments. They used different evaluation data, different features, and different approaches. In particular, they used data provided by the Document Understanding Conferences (DUC), and their annotation approach used two thresholds. They positively annotated the sentences with ROUGE score higher than 0.7 and negatively annotated those with score lesser than 0.3. Apparently, sentences with score between 0.3 and 0.7 were not used in their experiments.

We therefore replicated their annotation approach using the BioASQ data set and our features so that we could compare with our other experiments and obtained an average ROUGE-S4 of 0.09. This is lower than the results of our regression approach. 


Our results are therefore compatible with the results provided by \citet{Ouyang:2011} when we use their annotation approach for classification. We can consequently conclude that classification can deliver better results than regression, but we need to be careful with the approach used to annotate the training sentences.

\begin{figure}
\centering
\includegraphics[width=\linewidth]{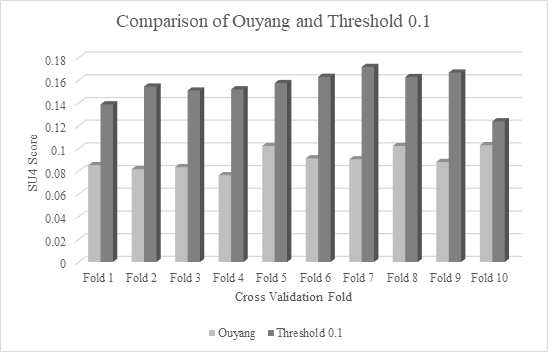}
\caption{Classification with Ouyang et al. and our annotation approach (0.1 as threshold).
\label{fig:ouyang}}
\end{figure}

Figure~\ref{fig:ouyang} provides a comparison of our best performing annotation with \citet{Ouyang:2011}'s approach by showing the variation of SU4 over all cross-validation folds.

The results reported in this paper are not directly comparable with the official results of the BioASQ runs for two reasons. First of all, the system implemented in this paper uses the entire source summaries as input. In contrast, systems participating in BioASQ can use additional information about what snippets from the source summaries are most relevant. Second, \citet{Molla2017} observed that the results of cross-validation with the training data gave much poorer results than the results evaluated using the BioASQ test set and the BioASQ evaluation scripts. Of the runs submitted by \citet{Molla2017}, only the one labelled RNN used as input the full summaries without information about relevant snippets. The average of ROUGE-SU4 across all batches was 0.435. However, our (unpublished) experiments revealed that cross-validation of the same system achieved a ROUGE-SU4 of 0.144. This is lower than our best results using classification reported in this paper.

\section{Conclusions}\label{sec:conclusions}

We have presented a comparison of two supervised machine learning techniques for extractive query focused summarisation. In addition, we have also explored the difficult phase of annotating data for classification approaches for summarisation, drawing a comparison among several annotation techniques.

To evaluate the model for both approaches, we have conducted an automatic evaluation and compared the performance of our system against human generated systems by using ROUGE. A series of experiments have been conducted by labelling data by different mechanisms for classification-based approaches. 

Our experiments revealed that classification performs better than regression when a threshold of 0.1 SU4 is applied for annotating data.

When comparing the different annotation techniques for the classification approach, we observed a considerable difference between the results when using threshold 0.1, using the highest three SU4 scoring sentences, or using other annotation techniques such as the ones by \citet{Marcu:1999} and \citet{Ouyang:2011}. 

As part of future work, we plan to conduct further experiments to determine the best annotation techniques for classification-based approaches. In particular, we plan to explore the impact of the second part of Marcu's greedy approach to see any improvement in results, along with utilising ROUGE as a similarity measure instead of cosine similarity to generate the extract. In addition, we will explore automatic approaches to determine the best thresholds. We empirically tried several thresholds and observed that 0.1 improved results but ideally this part would be done automatically.

We also plan to conduct an analysis of experiments by using learning to rank approaches. This type of learning algorithms may help improve performance.

\section*{Acknowledgements}

This research is partly funded by CSIRO's Data61.


\bibliography{louhi2018}
\bibliographystyle{acl_natbib_nourl}

\end{document}